\documentclass[sigconf,nonacm]{acmart}

\usepackage{multirow}
\usepackage{subcaption}
\usepackage{graphicx}
\usepackage{tikz}
\usepackage{comment}
\usepackage{algorithm}
\usepackage{algpseudocode}
\usepackage{booktabs}
\usepackage{tabularx}
\usepackage{graphicx}

\AtBeginDocument{%
  }

\setcopyright{acmlicensed}

\copyrightyear{2026}
\acmYear{2026}
\setcopyright{cc}
\setcctype{by}
\acmConference[WWW '26] {Proceedings of the ACM Web Conference 2026}{April 13--17, 2026}{Dubai, United Arab Emirates.}
\acmBooktitle{Proceedings of the ACM Web Conference 2026 (WWW '26), April 13--17, 2026, Dubai, United Arab Emirates}
\acmISBN{979-8-4007-2307-0/2026/04}
\acmDOI{10.1145/3774904.3793020}
\begin{document}

\title[Invisible Walls in Cities]{Invisible Walls in Cities: Designing LLM Agent to Predict Urban Segregation Experience with Social Media Content}

\author{Bingbing Fan}
\authornote{Both authors contributed equally to this research.}
\orcid{0009-0005-3671-056X}
\affiliation{%
  \institution{Department of Electronic Engineering, BNRist, Tsinghua University}
  \city{Beijing}
  \country{China}
}
\email{fbb24@mails.tsinghua.edu.cn}

\author{Lin Chen}
\authornotemark[1]
\orcid{0000-0002-2605-749X}
\affiliation{%
  \institution{Hong Kong University of Science and Technology}
  \city{Hong Kong}
  \country{China}}
\email{lchencu@connect.ust.hk}

\author{Songwei Li}
\orcid{0009-0001-1091-934X}
\affiliation{%
  \institution{Department of Electronic Engineering, BNRist, Tsinghua University}
  \city{Beijing}
  \country{China}
}
\email{lisw21@mails.tsinghua.edu.cn}

\author{Jian Yuan}
\orcid{0000-0001-9734-6056}
\affiliation{%
\institution{Department of Electronic Engineering, BNRist, Tsinghua University}
  \city{Beijing}
  \country{China}
  }
\email{jyuan@tsinghua.edu.cn}

\author{Fengli Xu}
\authornote{Corresponding author.}
\orcid{0000-0002-5720-4026}
\affiliation{%
  \institution{Department of Electronic Engineering, BNRist, Tsinghua University}
  \city{Beijing}
  \country{China}}
\email{fenglixu@tsinghua.edu.cn}

\author{Pan Hui}
\orcid{0000-0001-6026-1083}
\authornotemark[2]
\affiliation{%
  \institution{Hong Kong University of Science and Technology (Guangzhou)}
  \city{Guangzhou}
  \country{China}}
\affiliation{%
  \institution{Hong Kong University of Science and Technology}
  \city{Hong Kong}
  \country{China}}
\email{panhui@ust.hk}

\author{Yong Li}
\orcid{0000-0001-5617-1659}
\authornotemark[2]
\affiliation{%
\institution{Department of Electronic Engineering, BNRist, Tsinghua University}
  \city{Beijing}
  \country{China}}
\email{liyong07@tsinghua.edu.cn}

\renewcommand{\shortauthors}{Bingbing Fan et al.}

\begin{abstract}
  Understanding experienced segregation in urban daily life is crucial for addressing societal inequalities and fostering inclusivity. The abundance of user-generated reviews on social media encapsulates nuanced perceptions and feelings associated with different places, offering rich insights into segregation. However, leveraging this data poses significant challenges due to its vast volume, ambiguity, and confluence of diverse perspectives. To tackle these challenges, we propose a novel Large Language Model (LLM) agent to automate online review mining for segregation prediction. Specifically, we propose a \textit{reflective LLM coder} to digest social media content into insights consistent with real-world feedback, and eventually produce a codebook capturing key dimensions that signal segregation experience, such as \textit{cultural resonance and appeal}, \textit{accessibility and convenience}, and \textit{community engagement and local involvement}. Guided by the codebook, LLMs can generate both informative review summaries and ratings for segregation prediction. Moreover, we design a \textbf{\underline{RE}asoning-and-\underline{EM}bedding (RE'EM)} framework, which combines the reasoning and embedding capabilities of language models to integrate multi-channel features for segregation prediction. 
  Experiments on real-world data demonstrate that our agent substantially improves prediction accuracy, with a 22.79\% elevation in R$^{2}$ and a 9.33\% reduction in MSE. The derived codebook is generalizable across three different cities, consistently improving prediction accuracy. Moreover, our user study confirms that the codebook-guided summaries provide cognitive gains for human participants in perceiving places of interest (POIs)' social inclusiveness. Our study marks an important step toward understanding implicit social barriers and inequalities, demonstrating the great potential of promoting social inclusiveness with Web technology.\footnote{\textcolor{red}{This paper has been accepted at The ACM Web Conference 2026. Please cite the
version appearing in the conference proceedings.}}
\end{abstract}




\maketitle
\clearpage
\section{Introduction}

The rapid advancement of web technologies has enabled fine-grained records of social interactions through diverse media, such as textual posts~\cite{weller2014twitter}, photos~\cite{gilbert2013need}, and short videos~\cite{chen2024shorter}. 
As a result, web platforms have become extensive social sensing systems, generating rich digital traces that reflect social dynamics in the physical world~\cite{stier2021evidence,iqbal2023lady}. 
Such data offers a unique opportunity to uncover subtle patterns or preferences in individuals' everyday life, allowing researchers to study long-standing social issues from a new perspective~\cite{moro2021mobility}.
Among the most enduring of these issues is segregation: the systematic separation of individuals or groups based on certain characteristics.
However, identifying cues of segregation experience from massive, multimodal web content poses significant challenges, requiring advanced methods to capture nuanced feelings, and there is no established procedure to guide this process. 
Historically, research on segregation has focused primarily on residential patterns~\cite{park25city}, where individuals of the same racial or ethnic group are more likely to reside in the same neighborhoods~\cite{schelling2006micromotives}. 
This type of spatial segregation has been linked to negative social outcomes, such as limited upward social mobility and increased crime rates~\cite{gordon2006urban,kramer2009segregation}.
Recent studies have shifted focus to experienced segregation—the dynamic segregation individuals experience in daily movements~\cite{moro2021mobility}.
Despite the seemingly free human movements in most modern urban spaces, certain demographic groups continue to face barriers to equal access and social interactions.
In other words, there seem to be “invisible walls” in cities that prevent sufficient social interactions between groups. 
These walls are not merely the result of physical proximity but are influenced by cultural, social, and economic factors that drive segregation on a more nuanced level.
Accurately predicting segregation experiences can inform individuals to avoid potentially uncomfortable situations and support policymakers in fostering social inclusiveness.
Moreover, research suggests that segregation experiences can intensify in larger urban areas~\cite{nilforoshan2023human}, underscoring the need to explore the underlying mechanisms for growing more equitable cities admist ever-increasing urbanization.
However, existing studies only present retrospective and holistic measurements, providing limited insights and prediction capabilities.

To address these challenges, we propose to leverage the reasoning power of LLMs for automated social media content analysis. 
We propose a \textit{reflective LLM coder}, featuring a strategic agentic workflow that integrates two key components: a \textit{reflective attributor} and a \textit{code summarizer}. 
The \textit{reflective attributor} employs an abductive reasoning approach: 
it prompts the LLM to estimate a location’s appeal to different demographic groups from its social media content, and reconcile discrepancies through iterative reflection on observed segregation patterns. 
Subsequently, the \textit{code summarizer} applies chain-of-thought reasoning~\cite{wei2022chain} to iteratively merge insights into a structured codebook. 
This codebook guides LLMs to transform free-text reviews into structured summaries, highlighting segregation-related factors like cultural resonance and community engagement.
Complementing this qualitative approach, we propose a \textbf{\underline{RE}asoning and \underline{EM}bedding (RE'EM) framework} for quantitative segregation prediction, which combines LLMs' reasoning capabilities with the representational power of pre-trained embedding models. 
First, we prompt the LLM to provide structured ratings of a place’s appeal to different groups based on the codebook, allowing the reasoning outcomes of LLMs to be vectorized and easily integrated with other channels of features. 
Concurrently, we finetune an embedding model to learn global representations optimized for segregation prediction. 
Finally, we fuse structured ratings, global embeddings, and population information using a neighbor-aware multi-view predictor. 

We validate our approach through qualitative user studies and quantitative experiments on four US cities. 
In our user study with 75 researchers, participants' prediction accuracy greatly improved when provided with LLM-generated summaries.
Furthermore, as many as 80\% of participants prefer the codebook-guided summaries over vanilla LLM outputs. 
In quantitative experiments, the RE'EM framework improves the predictive $R^2$ by 22.79\% and reduces the MSE by 9.33\% compared to baseline models relying solely on local racial composition, and is generalizable across cities.

Our contributions are summarized as follows:
\begin{itemize}
\item We are the first to explore the use of social media content for predicting POI experienced segregation.
\item We design a reflective LLM coder to effectively summarize online reviews and identify cues of segregation.
\item We propose a REasoning-and-EMbedding (RE'EM) framework that combines the reasoning and embedding capabilities of language models to predict segregation.
\item We validate our approach through comprehensive qualitative and quantitative evaluations, demonstrating its effectiveness in extracting operationable insights and generalizing segregation prediction improvement. 
\end{itemize}







\section{Related Work}

\noindent \textbf{Mining Web Data with LLMs.}
The digital footprints and rich textual contents people generate on Web platforms have proven to be informative for a wide array of social phenomena, including health outcome~\cite{nguyen2016building}, mental well-being~\cite{mitchell2013geography}, crime rates~\cite{fatehkia2019correlated}, and neighborhood disparities~\cite{rama2020facebook,iqbal2023lady}. However, these studies often rely on pre-calculated indices or pre-defined word lists to engineer and extract features.
This is not only labor-intensive but also lacks adaptability to evolving research questions and contexts.
Given the remarkable language understanding and reasoning capabilities of LLMs, recent research explored the potential of leveraging LLMs to mine Web data for social good, such as revealing food-related social prejudice~\cite{luo2024othering}, evaluating public accessibility~\cite{li2024toward}, and capturing urban perception~\cite{santos2024real}.
Nevertheless, these studies often lack an examined framework for insight extraction from massive social media content, which limits their effectiveness in capturing the complexity of Web data.
In contrast, our work presents the first step toward unlocking the reasoning power of LLMs to code free-form online reviews, extracting human-comprehensible, informative insights for segregation prediction.

\noindent \textbf{Experienced Segregation.}
The study of segregation traces its roots back to the early 20th century, when sociologists examined the division of urban spaces into ``ecological niches'', each inhabited by distinct social groups~\cite{park25city}. 
Researchers revealed the detrimental effects of enduring segregation even in the absence of ``legalized racial segregation'', on education, income, housing, and crime~\cite{king1973racial,massey1987effect,charles2003dynamics}.
Albeit offering valuable insights, these studies are highly constrained by data availability, overlooking the reality that individuals spend significant time and engage in numerous interactions beyond the confines of their residential neighborhoods~\cite{wang2018urban}.
With the rapid proliferation of smart mobile devices, the ability to track people's intricate movements in urban spaces has emerged~\cite{xu2016understanding}.
This has paved the way for a novel research avenue aimed at quantifying and understanding the type of segregation individuals actually \textit{experience} in their daily movements~\cite{moro2021mobility,athey2021estimating,de2024people}, including how it may change during disasters~\cite{yabe2023behavioral,chen2023getting} and scale with city sizes~\cite{nilforoshan2023human}.
Nevertheless, existing studies primarily focus on static spatial distribution of demographic groups and physical movement, overlooking the complex socioeconomic factors behind the segregation phenomenon, such as cultural resonance and local involvement.
Thus, these works can only provide retrospective measurement but have limited predictive power. 
In contrast, our work establishes an LLM-based method to automatically extract nuanced features from online reviews for experienced segregation prediction.
\begin{figure*}[t]
    \centering
    \includegraphics[width=0.99\linewidth]{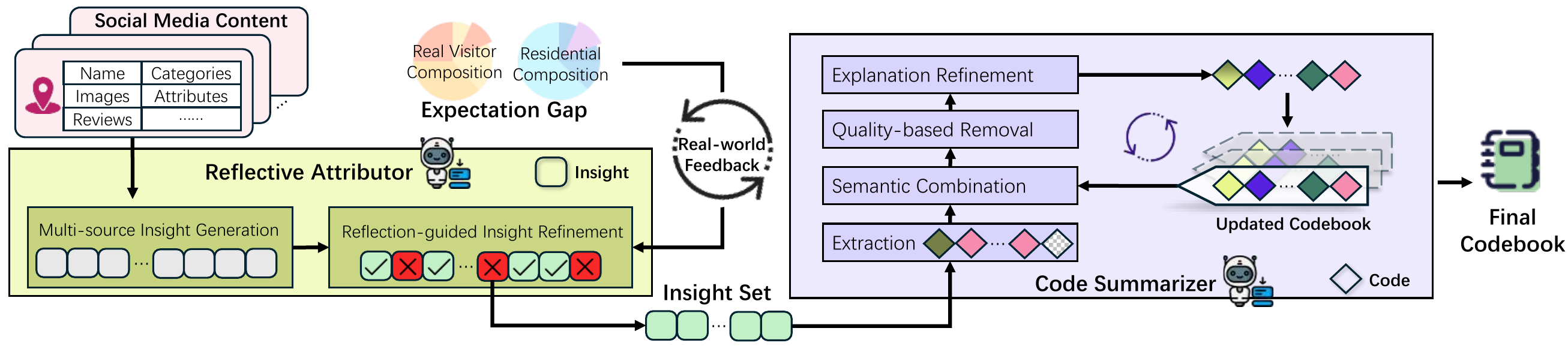}
    \caption{Overview of the \textit{reflective LLM coder}. It consists of a \textit{reflective attributor} that integrates multi-source review and image signals and refines insights using real visitation patterns, and a \textit{code summarizer} that consolidates these insights into a structured, generalizable codebook capturing key factors shaping experienced segregation.}
    \label{fig:coding}
\end{figure*}

\section{Preliminary}

\subsection{Problem Formulation}

Inspired by~\cite{moro2021mobility}, we compute the proportion of visitors at POI $i$ by each racial group $q$, denoted as $\tau_{qi}$. We then quantify the segregation at POI $i$ as the deviation of the visitor proportion $\tau_{qi}$ from the city's residential proportion $T_q$ (k is a constant that normalizes $S_{i}$ to range between 0 and 1):
\begin{equation}
    S_{i} = k\sum_{q}(|\tau_{iq}-T_{q}|).
\end{equation}

Our task can be formulated into a POI segregation prediction problem.
Given a set of POIs $I = \{i_1, i_2, ..., i_N\}$ along with their corresponding social media content $C = \{c_1, c_2, ..., c_N\}$
and the racial composition of local population $P = \{p_1, p_2, ..., p_N\}$, the objective is to learn a function $f(i, c, p)$ to minimize the discrepancy between the predicted segregation and the real segregation $S_{i}$:
\begin{equation}
    min~\text{diff}(f(i, c, p), S_{i}). 
\end{equation}

\subsection{Data}

We use three datasets to obtain social media, demographic, and mobility information. 
Social media and demographic data serve as input features, while mobility data is used to calculate the ground truth segregation at POIs, i.e., our prediction targets.
We select cities based on the sufficiency of available POIs within the overlapping scope of the three datasets.
As a result, we identify four cities for analysis: Philadelphia (5,360 POIs), Tucson (2,703 POIs), Tampa (2,222 POIs), and New Orleans (2,392 POIs). 

\textbf{Demographic data.}
We obtain the demographics of census block group (CBG) residents from the American Community Survey 5-year Estimates (ACS)\footnote{https://www.census.gov/programs-surveys/acs/}.
A CBG is the smallest spatial unit with available demographics, typically consisting of 600-3,000 residents.
We categorize CBG residents into five distinct groups: \textit{hispanic}, \textit{black}, \textit{asian}, \textit{white}, and \textit{others}, and calculate their respective ratios by dividing the population of each group by the CBG population. 

\textbf{Social media data.}
We use the Yelp Open Dataset\footnote{https://www.yelp.com/dataset} for social media data. 
For each POI, we obtain its name, location, stars, categories, service attributes, and multi-modal user-generated content. 
The category captures its business domain, such as ``food'' and ``department stores''. 
The attributes reflect its specific business traits, such as price range, delivery options, and parking availability.
The user-generated content includes textual reviews and images uploaded by visitors. 
Each POI is associated with at least 5 reviews. 

\textbf{Mobility data.}
To derive the ground truth of experienced segregation, we utilize the Safegraph Patterns Dataset (accessed through Advan)\footnote{https://www.deweydata.io/data-partners/advan}.
This dataset records monthly visits to each POI originating from various CBGs.
We aggregate one year's visitation records to construct a robust dataset with extensive observations.
To address the distinct POI indexing systems between Yelp and SafeGraph, we match POIs by ensuring identical names and a location deviation within 200 meters. 
In Appendix~\ref{appendix:a}, we show the variability of POI features, highlighting the importance of our study.

We use demographic and mobility data from 2019 for temporal consistency. To leverage historical POI information, we include social media content from 2019 (16\%) and the preceding eight years (84\%). Each review is timestamped, enabling LLM to reason about the temporal relationship between the content and the current analysis period.



\section{Reflective LLM Coder}
The vast volume of raw social media content creates a significant cognitive barrier to revealing the encoded segregation experience. 
Therefore, an effective approach is needed to process and extract insights.
Building on research demonstrating the consistency and stability of LLM-based coding~\cite{dai2023llm,tai2024examination}, we design a \textit{reflective LLM coder} to generate an insightful codebook from a small set of POI data, capturing the multi-faceted factors influencing segregation and offering a structured framework for deeper analysis.
As shown in Figure~\ref{fig:coding}, it contains two modules, \textit{reflective attributor} and \textit{code summarizer}.

\subsection{Reflective Attributor}

We design a \textit{reflective attributor} to identify factors influencing a POI’s appeal to different groups. 
Specifically, we design a two-step Chain-of-Thought (CoT) scheme~\cite{wei2022chain} that guides an LLM to integrate multiple perspectives from texts and images, forming a reflection-driven process that refines insights based on real-world feedback.

\textbf{Multi-source insight generation.}
In this step, the LLM generates insights by integrating multiple sources of information: the POI's name, user-posted reviews, and images. 
It first evaluates the name for potential cultural, racial, ethnic, or socioeconomic associations that might influence the appeal to specific groups. 
Next, it analyzes the reviews and images to identify key aspects like atmosphere, pricing, service quality, and cultural relevance. 
These analyses are then synthesized into a concise set of insights, each identifying factors that could attract or repel particular racial/ethnic groups, ensuring comprehensive coverage without redundancy.




\begin{table*}[ht]
\centering
\caption{Automatically constructed codebook. Each dimension represents a distinct driver of POI attractiveness or deterrence for different demographic groups, forming the conceptual “bricks” underlying experienced segregation patterns. The explanations summarize the thematic meaning of each code.}
\vspace{-10pt}
\scalebox{0.85}{
\begin{tabular}{@{}cp{4.2cm}p{15.2cm}@{}}
\toprule
\textbf{Index} & \textbf{Name} & \textbf{Detail} \\ 
\midrule
1  & \textbf{Cultural Resonance and Appeal}  & Culturally themed offerings, such as Italian-American or South Indian cuisine, attract visitors seeking authentic or familiar experiences, influencing visitation based on cultural representation and resonance. \\
\midrule
2  & \textbf{Price Sensitivity and Economic Accessibility}  & Moderate pricing, coupons, and cost-effective policies like BYOB appeal to budget-conscious visitors, impacting visitation patterns based on affordability and economic considerations.  \\
\midrule
3  & \textbf{Service Quality and Customer Experience}  & Professional and attentive service, despite occasional inconsistencies, attracts visitors valuing high service standards and personal interactions, influencing demographics based on service expectations.  \\
\midrule
4  & \textbf{Atmosphere and Social Environment}  & Lively, trendy, or family-friendly settings attract visitors prioritizing social and communal experiences, impacting visitation based on social and family-oriented preferences.  \\
\midrule
5  & \textbf{Accessibility and Convenience}  & Central locations, parking availability, and delivery services attract visitors prioritizing efficiency and accessibility, influencing patterns based on transportation and convenience.  \\
\midrule
6  & \textbf{Visual and Aesthetic Appeal}  & Modern, chic, and historically themed environments attract visitors who appreciate aesthetic and immersive experiences, influencing demographics based on visual and cultural preferences.  \\
\midrule
7  & \textbf{Cultural and Social Inclusivity}  & Inclusive, diverse, and culturally sensitive environments attract a broad demographic by catering to varied identities and preferences, influencing visitor composition based on inclusivity and cultural representation.  \\
\midrule
8  & \textbf{Product Variety and Quality}  & Diverse and high-quality offerings, including visually appealing and culturally themed products, attract visitors prioritizing variety and quality, influencing visitation based on product expectations. \\
\midrule
9  & \textbf{Community Engagement and Local Involvement}  & Establishments with strong community ties and neighborhood vibes attract visitors valuing local engagement and communal experiences, influencing demographics based on community integration and involvement.  
 \\
\bottomrule
\end{tabular}
}
\label{tab:codebook}
\end{table*}

\textbf{Reflection-guided insight refinement.}
Importantly, we introduce real-world data to guide LLM's reflections upon these insights. 
We measure the diﬀerence between the demographic composition of visitors to each POI and that of nearby residents, using it as a ground-truth signal to identify which racial/ethnic groups it attracts or repels. 
This signal guides the LLM in refining the existing insights in an abductive manner, retaining only those consistent with real-world evidence.
This step not only enhances the relevance of the insights, but also provides a grounded explanation for observed visitation preferences.

\begin{figure*}[t]
    \centering
    \includegraphics[width=0.95\linewidth]{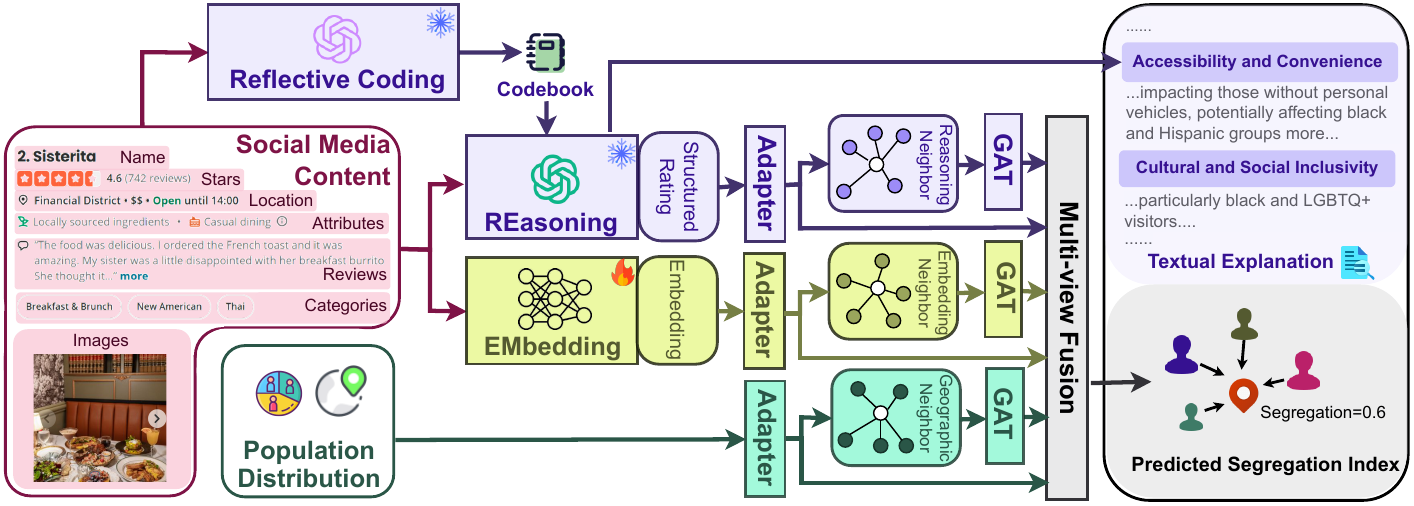}
    \caption{Reasoning-and-Embedding (RE'EM) framework. RE'EM integrates three complementary channels (reasoning, embedding, and population) with a neighbor-aware multi-view fusion to predict POIs' experienced segregation.}
    \label{fig:reem}
\end{figure*}

Combining these two steps, the \textit{reflective attributor} links POI attributes with patterns in visitation behavior, offering a data-driven understanding of segregation experiences.

\subsection{Code Summarizer}

The \textit{code summarizer} module aggregates outputs from the \textit{reflective attributor} into a comprehensive and orthogonal codebook that encapsulates the diverse factors influencing POI appeal. 
We design another four-step CoT scheme to guide the LLM reasoning process.

\textbf{Extraction.} 
Taking the insights extracted by the \textit{reflective attributor}, this step extracts phrases representing the analysis perspectives for each insight. 
It ensures that previously-identified features are captured for further refinement.

\textbf{Semantic combination.} 
This step combines insights with similar semantics to reduce redundancy, e.g., merging “Cultural Relevance and Appeal” and “Inclusivity and Cultural Representation” into a unified and coherent code. 

\textbf{Quality-based removal.}
This step filters out insights that are less distinguishable or less generalizable.
It ensures that the final codebook focuses on the most relevant and impactful factors, reducing the impact of noise. 

\textbf{Explanation refinement.} 
Finally, this step refines the one-sentence explanation for each code, highlighting broader, thematic insights to ensure flexibility and interpretability across contexts while preserving analytical rigor.


Trough the above coding process (formalized in Appendix ~\ref{appendix:b}), we obtain \textbf{a codebook identifying 9 distinct ``bricks'' that form the invisible walls}, shown in Table~\ref{tab:codebook}.
Each brick represents a key aspect of the visitor experience, allowing a nuanced understanding of how places appeal to diverse racial demographics.
For example, \textit{Cultural Resonance and Appeal} captures how culturally themed offerings such as Italian-American cuisines attract visitors seeking authentic experiences, while \textit{Price Sensitivity and Economic Accessibility} captures the impact of affordability on visitation, with moderate pricing and cost-effective policies like BYOB appealing to budget-conscious visitors.

\section{REasoning-and-EMbedding (RE'EM) Framework}

To effectively integrate different features for accurate segregation prediction, we design a \textbf{REasoning-and-EMbedding (RE’EM)} framework composed of four key components: a \textit{reasoning channel}, an \textit{embedding channel}, a \textit{population channel}, and a \textit{neighbor-aware multi-view predictor}, as shown in Figure~\ref{fig:reem}.

\subsection{Reasoning Channel}

We prompt an LLM to assess place attractiveness to different racial/ethnic groups under the guidance of our codebook (Table~\ref{tab:codebook}). 
Specifically, we instruct the LLM to simulate the perspective of each racial/ethnic group, and rate the POI along the codebook-identified dimensions. 
To unleash the LLM's reasoning power, we design a two-step CoT scheme.

\textbf{Codebook-guided summary.}
For each POI $i$, we denote its associated social media content as $c_i$, which includes the POI's name, review text, and images (if any). 
We instruct the LLM to analyze $c_i$ and generate a structured summary $u_i$ of the POI’s characteristics along the 9 codebook-defined dimensions, focusing on features that may attract or repel specific racial/ethnic groups.

\textbf{Structured rating.}
Based on the summary $u_i$, we ask the LLM to imagine itself as a member of each of the 5 racial/ethnic groups and assign ratings to the POI across the 9 dimensions. 
Ratings range from 0 (strong repulsion) to 10 (strong attraction), with 5 indicating neutrality.
This process yields a 45-dimensional vector $r_{i}$ with corresponding textual explanations $e_i$, providing both quantitative scores and qualitative insights into the POI’s social inclusiveness.
\begin{equation}
    r_i, e_i = \text{LLM}(u_{i}, \text{codebook}).
\end{equation}
To integrate the ratings into our predictive framework, we employ a \textit{reasoning adapter} consisting of a multi-layer perceptron (MLP). 
It transforms the ratings into a vector representation $v_{i}^{r}$:
\begin{equation}
    v_{i}^r = \text{ReasonAdapter}(r_i).
\end{equation}

\subsection{Embedding Channel}
The \textit{embedding channel} extracts deep semantic representations $v_{i}^{e}$ from the review corpus associated with each POI $i$. 
Due to the token length constraints, we apply data augmentation, i.e., sample different subsets of $c_i$ for input to the embedding model.
We fine-tune an open-source text embedding model, GTE-base~\cite{GTE-base}, to enhance its capability of extracting higher-level representations while preserving the pre-trained semantic knowledge. 
Although both the reasoning and embedding channels take $c_{i}$ as input, they extract complementary aspects: structured attribute reasoning in the former, and holistic semantic embedding in the latter.
The resulting embedding is then passed through an \textit{embedding adapter} (MLP) to produce $v_{i}^{e}$.
\begin{equation}
    v_{i}^e=\text{EmbeddingAdapter}(\text{GTE}(c_{i})).
\end{equation}

\subsection{Population Channel}
For each POI $i$, we compute a 5-dimensional feature vector $p_i$ representing the racial composition of the surrounding population. 
Specifically, we identify all CBGs whose centroids fall within a 0.5km radius around the POI, and compute a population-weighted average of their racial compositions.
The resulting $p_i$ reflects the static demographic context of the POI's geographic location.
We apply a \textit{population adapter} (another MLP) to transform $p_{i}$ into a latent representation $v_{i}^p$ for subsequent prediction.
\begin{equation}
    v_{i}^p=\text{PopulationAdapter}(p_{i}).
\end{equation}

\subsection{Neighbor-Aware Multi-View Predictor} 
This predictor primarily performs two key operations: neighbor aggregation and multi-view fusion.
For each POI $i$, we identify its five nearest neighbors in three separate spaces (views), respectively: the reasoning space (based on similarity to $r_{i}$), the embedding space (similarity to $e_{i}$), and the geographic space (physical proximity).
This forms three neighbor sets $N_{i}^r=\{n_1^r,...,n_5^r\}$, $N_{i}^e=\{n_1^e,...,n_5^e\}$ and $N_{i}^p=\{n_1^p,...,n_5^p\}$.
We use a graph attention network (GAT) for each view to aggregate the features of the respective neighbor set, producing neighbor-aware representations $v_{n_{i}}^r$, $v_{n_{i}}^e$, and $v_{n_{i}}^p$:
\begin{equation}
    v_{n_{i}}^*=\text{GAT}_{*}(v_{i}^*,\{v_{n}^*|n\in N_{i}^*\})
\end{equation}
Finally, a fully connected fusion module integrates both the POI's representations $v_{i}^*$ and its neighrbors' representations $v_{n_{i}}^*$ from all three channels.
This fusion process ultimately outputs the predicted segregation $\hat{S_{i}}$:
\begin{equation}
    \hat{S_{i}}=\text{Fuse}(v_{i}^r,v_{n_{i}}^r,v_{i}^e,v_{n_{i}}^e,v_{i}^p,v_{n_{i}}^p). 
\end{equation}


\begin{table*}[t]
\centering
\caption{Segregation prediction performances across different method. Compared with population-only and embedding baselines, RE’EM achieves significant improvements across MSE, RMSE, MAE, and R².}
\vspace{-10pt}
\scalebox{0.75}{
\begin{tabular}{@{}crrrrrrrr@{}}
\toprule
Model &
  \multicolumn{1}{c}{MSE↓} &
  \multicolumn{1}{c}{Improv.} &
  \multicolumn{1}{c}{RMSE↓} &
  \multicolumn{1}{c}{Improv.} &
  \multicolumn{1}{c}{MAE↓} &
  \multicolumn{1}{c}{Improv.} &
  \multicolumn{1}{c}{R$^2$↑} &
  \multicolumn{1}{c}{Improv.} \\ \midrule
Population & 0.0075±0.0002 & ------ & 0.0864±0.0012 & ------ & 0.0683±0.0013 & ------  & 0.3164±0.0204 & ------ \\
GTE        & 0.0073±0.0002 & 2.67\% (p=0.7623) & 0.0855±0.0011 & 1.04\% (p=0.7887) & 0.0685±0.0010 & -0.29\% (p=0.9713) & 0.3249±0.0151 & 2.69\% (p=0.8905) \\
BERT       & 0.0072±0.0002 & 4.00\% (p=0.0147) & 0.0846±0.0011 & 2.08\% (p=0.0181) & 0.0680±0.0010 & 0.44\% (p=0.3708)  & 0.3394±0.0176 & 6.78\%(p=0.0549) \\
GloVE      & 0.0072±0.0002 & 4.00\% (p=0.0242) & 0.0848±0.0011 & 1.85\% (p=0.0290) & 0.0685±0.0008 & -0.29\% (p=0.6118) & 0.3357±0.0140 & 5.75\% (p=0.0776) \\
Qwen3-Embedding-8B & 0.0072±0.0002 & 4.00\% (p=0.0196) &0.0848±0.0011 & 1.85\% (p=0.0292) & 0.0683±0.0009 & 0.00\% (p=0.4789) & 0.3349±0.0181 & 5.85\% (p=0.0812) \\
\textbf{RE`EM}&
  \textbf{0.0068±0.0003} &
  \textbf{9.33\% (p=0.0080)} &
  \textbf{0.0823±0.0021} &
  \textbf{4.75\% (p=0.0071)} &
  \textbf{0.0662±0.0021} &
  \textbf{3.07\% (p=0.0597)} &
  \textbf{0.3885±0.0278} &
  \textbf{22.79\% (p=0.0022)} \\ \bottomrule
\end{tabular}
}
\label{tab:prediction_result}
\end{table*}
\section{User Study}
We conduct a user study to evaluate the cognitive gain to humans provided by our generated codebook.
We recruit participants via snowball sampling among graduate students and researchers with hands-on experience or sufficient knowledge in social media studies: approximately 17.3\% from urban planning, 24\% from social science, and 58.7\% from data mining.
To ensure that participants fully understood the survey tasks, we provided clear explanations of the geographic and racial concepts required for the study.
Our questionnaire is included in our released code repository \footnote{https://github.com/tsinghua-fib-lab/Invisible-Wall-in-Cities}.

We randomly sample four POIs whose visitor demographics significantly differ from local residents. 
For each POI, we generate review summaries using a vanilla method (directly prompting LLMs without the codebook) and our codebook-guided method (see Appendix Figure~\ref{fig:case-study}).
Participants answer four questions per POI, progressively incorporating different levels of contextual information.
(1) \textbf{Baseline estimation}: Participants predict whether each racial group is over- or underrepresented at the POI based solely on its basic attributes (e.g., name, category, and service details).
(2) \textbf{Review-informed prediction}: Participants reassess their estimates after being provided with a random sample of user reviews.
(3) \textbf{Summary-enhanced prediction}: Participants make a final prediction after reviewing a summary of the user reviews.
(4) \textbf{Summary preference evaluation}: Participants determine whether the codebook-guided summary or the vanilla summary is more informative.

\begin{figure}[t]
    \centering
    \includegraphics[width=0.9\linewidth]{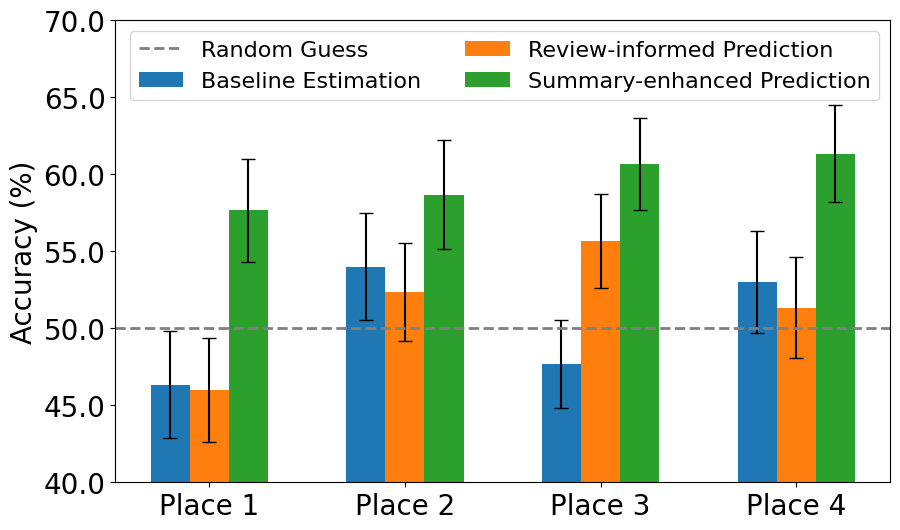}
    \caption{Human prediction accuracy with different information availability. Accuracy remains near random (50\%) when only POI metadata (\textbf{Baseline Estimation}) or raw reviews (\textbf{Review-informed Prediction}) are available, but increases substantially once participants receive codebook-guided LLM summaries (\textbf{Summary-enhanced Prediction}), demonstrating the cognitive benefit of structured review distillation.}
    \label{fig:user-study-1}
\end{figure}

\begin{figure}[t]
    \centering
    \includegraphics[width=0.9\linewidth]{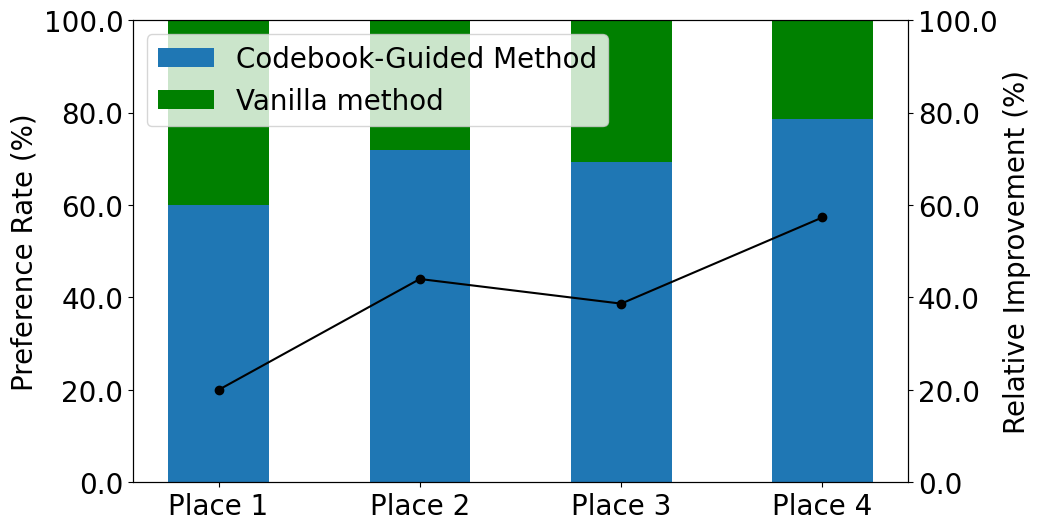}
    \caption{User preference between codebook-guided summary and vanilla summary. Across all sampled POIs, a strong majority of participants favor codebook-guided summaries, indicating that structured summaries better support human understanding of POI inclusiveness. }
    \label{fig:user-study-2}
\end{figure}

Figure~\ref{fig:user-study-1} presents the human prediction accuracies with different information availability. 
When relying solely on basic POI attributes, accuracy fluctuates around 50\%, indicating that visitor preferences cannot be reliably inferred from POI metadata alone. 
Moreover, introducing randomly sampled user reviews does not lead to a notable improvement in prediction accuracy.
This is likely due to the sparsity and unstructured nature of raw review data, which can be difficult to process and may even mislead human judgment. 
In contrast, providing participants with the review summary consistently enhances prediction accuracy across all sampled POIs.
This result underscores the effectiveness of the codebook-guided summarization, which condenses extensive social media content into concise yet highly informative text snippets, enhancing human understanding of POIs' social inclusiveness.

Figure~\ref{fig:user-study-2} presents the results of the summary preference evaluation.
Across all sampled POIs, 60\%-80\% of participants prefer the codebook-guided summary over the vanilla one, with an average improvement of 40\%, showing that our approach more effectively distills useful information to support human judgment.
As Appendix ~\ref{appendix:d} illustrates, the codebook-guided summaries capture nuanced insights on the potential deterrants for certain minority groups, which are often absent or less explicit without codebook guidance. 

\section{Prediction Experiments}

\subsection{Experiment Settings}

We compare our model against five baselines. 
The first baseline is an MLP making use of only the population information.
The other four baselines process the social media content with a frozen pre-trained text embedding model, and fuse it with population information.
We adopt four widely-used powerful embedding models: GTE-base~\cite{GTE-base}, BERT-base~\cite{BERT}, GloVE-330B~\cite{glove} and Qwen3-Embedding-8B~\cite{qwen3embedding}. 
The structures of the population and embedding MLPs are identical to those in our model.
We use four metrics to evaluate model performances: mean squared error (MSE), root mean squared error (RMSE), mean absolute error (MAE), and coefficient of determination (R\textsuperscript{2}).

We employ GPT-4o for reflective coding on a small subset (n=190) of POIs, and GPT-4o-mini for codebook-guided reasoning on all POIs for economic considerations.
All models are trained using the Adam optimizer to minimize the MSE loss, with hyperparameters (e.g., learning rate, weight decay) tuned based on validation performances. 
POIs in each city are randomly split into training, validation, and test sets with a ratio of 6:2:2.
For further implementation details, please refer to Appendix~\ref{appendix:c} and our released code.
All LLM prompts used in this work are also provided in the repository.

\begin{table*}[ht]
    \centering
    \caption{Ablation study of RE'EM components. Using a single channel (row 1-3) or removing the codebook (row 4) leads to performance degradation. The full model consistently achieves the best results, highlighting the complementary contributions of reasoning, embedding, and population features.}
    \vspace{-10pt}
    \scalebox{0.99}{
    \begin{tabular}{@{}ccccc@{}}
    \toprule
     Model     & MSE↓          & RMSE↓        & MAE↓           & R$^2$↑             \\ \midrule
    Population & 0.0075±0.0002             & 0.0864±0.0012             & 0.0683±0.0013             & 0.3164±0.0204               \\
     Embedding & 0.0096±0.0002             & 0.0980±0.0011             & 0.0814±0.0008            & 0.1118±0.0052              \\
    
    Rating & 0.0104±0.0003            & 0.1021±0.0013             & 0.0837±0.0010            & 0.0384±0.0117              \\
    w/o Codebook & 0.0072±0.0002 & 0.0849±0.0016 &0.0685±0.0018 & 0.3499±0.0219 \\

   \textbf{Full Model} & \textbf{0.0068±0.0002}  & \textbf{0.0823±0.0021} &  \textbf{0.0662±0.0021}  & \textbf{0.3885±0.0278}   \\ \bottomrule
    \end{tabular}
    }
    \label{tab:Ablation}
\end{table*}

\begin{table*}[ht]
\centering
\caption{Cross-city generalization performances using codebook derived from Philadelphia. RE’EM outperforms all baselines across Tucson, Tampa, and New Orleans, demonstrating strong transferability.}
\vspace{-10pt}
\scalebox{0.99}{
\begin{tabular}{@{}cccrrrr@{}}
\toprule
City &
  POI Num &
  Model &
  \multicolumn{1}{c}{MSE↓} &
  \multicolumn{1}{c}{RMSE↓} &
  \multicolumn{1}{c}{MAE↓} &
  \multicolumn{1}{c}{R$^2$↑}  \\ \midrule
\multirow{5}{*}{Tucson} &
  \multirow{5}{*}{2,221} &
  GTE &
  0.0048±0.0001 &
  0.0692±0.0011 &
  0.0565±0.0009 &
  0.2117±0.0134  \\
 &
   &
  BERT &
  0.0049±0.0002 &
  0.0698±0.0012 &
  0.0572±0.0011 &
  0.1983±0.0165 \\
 &
   &
  GloVE &
  0.0047±0.0002 &
  0.0684±0.0011 &
  0.0560±0.0011 &
  0.2287±0.0175 \\
 &
 &
  Qwen3-Embedding-8B &
  0.0042±0.0001 &
  0.0645±0.0012 &
 0.0532±0.0010 &
  0.3147±0.0244 \\
 &
   &
  \textbf{RE'EM} &
  \textbf{0.0039±0.0001} &
  \textbf{0.0625±0.0009} &
  \textbf{0.0506±0.0011} &
  \textbf{0.3744±0.0290} \\ \midrule
\multirow{5}{*}{Tampa} &
  \multirow{5}{*}{2,703} &
  GTE &
  0.0049±0.0003 &
  0.0697±0.0018 &
  0.0578±0.0018 &
  0.2905±0.0300  \\
 &
   &
  BERT &
  0.0052±0.0002 &
  0.0719±0.0016 &
  0.0594±0.0016 &
  0.2455±0.0255  \\
 &
   &
  GloVE &
  0.0049±0.0002 &
  0.0704±0.0015 &
  0.0582±0.0016 &
  0.2762±0.0268 \\
 &
 &
  Qwen3-Embedding-8B&
  0.0049±0.0002 &
  0.0702±0.0017 &
  0.0581±0.0017 &
  0.2800±0.0240 \\
 &
   &
  \textbf{RE'EM} &
  \textbf{0.0044±0.0002} &
  \textbf{0.0668±0.0014} &
  \textbf{0.0525±0.0017} &
  \textbf{0.3499±0.0224}  \\ \midrule
\multirow{5}{*}{New Orleans} &
  \multirow{5}{*}{2,391} &
  GTE &
  0.0100±0.0006 &
  0.1002±0.0028 &
  0.0814±0.0029 &
  0.1958±0.0135 \\
 &
   &
  BERT &
  0.0110±0.0006 &
  0.1051±0.0029 &
  0.0858±0.0030 &
  0.1154±0.0101 \\
 &
   &
  GloVE &
  0.0099±0.0006 &
  0.0996±0.0029 &
  0.0804±0.0029 &
  0.2045±0.0209 \\
 &
 &
  Qwen3-Embedding-8B &
  0.0103±0.0005 &
  0.1013±0.0028&
  0.0827±0.0029 &
  0.1771±0.0106 \\
 &
   &
  \textbf{RE'EM}  &
  \textbf{0.0087±0.0005} &
  \textbf{0.0929±0.0028} &
  \textbf{0.0745±0.0024} &
  \textbf{0.2781±0.0424}  \\ \bottomrule
\end{tabular}}
\label{tab:prediction_result city generalize}
\end{table*}

\subsection{Experiment Results}
Table~\ref{tab:prediction_result} presents the performances of our RE'EM model alongside four baselines, with results reported as the mean and standard deviation over five repetitions on 50\% randomly sampled test sets. 
A key insight is that \textbf{all models incorporating social media data outperform the population-only baseline}. 
This result underscores the predictive power of social media content, which provides crucial insights beyond static population distributions. 
It supports our hypothesis that segregation patterns are shaped not only by demographic distributions but also by the social and cultural perceptions reinforcing the “invisible walls” in cities. 
Further, \textbf{RE'EM consistently and significantly surpasses all embedding-based baselines}.
It achieves a 9.33\%, 4.75\%, and 3.07\% improvement in MSE, RMSE, and MAE, respectively, over the population-only baseline. 
R\textsuperscript{2} increases to 0.3885, marking a substantial 22.79\% improvement.
These results indicate that while social media data encapsulates rich and valuable information, a naive embedding strategy is insufficient to fully capture the interplay between population distribution, POI attributes, and visitor dynamics. 
In contrast, the structured multi-view approach of RE'EM effectively integrates heterogeneous information sources, enabling accurate predictions of segregation.
We provide further analyses in Appendix ~\ref{appendix:additional}, showing that RE’EM remains stable across different base LLMs, and direct LLM prompting performs substantially worse than RE`EM architecture.

To validate our design of different model components, we perform an ablation study. 
As Table~\ref{tab:Ablation} shows, each channel (reasoning, embedding, population) exhibits certain predictive capabilities, but none achieves optimal performance. 
Without the codebook, relying solely on vanilla LLM rating yields markedly worse performance.
The full RE'EM model, which strategically combines all three channels and leverages the codebook for structured reasoning, consistently delivers superior results. 
Thus, the integration of multi-view signals enables complementary information flow, allowing the model to refine its predictions beyond what any single feature source can achieve.

To assess the generalization capability across cities, we use the codebook obtained in Philadelphia to train models in the other three cities. 
As shown in Table~\ref{tab:prediction_result city generalize}, RE'EM consistently achieves the best performance in all three cities along all metrics. 
Compared to the strongest baseline, RE'EM achieves substantial improvements, with MSE reduced by as much as 12.12\%, MAE reduced by as much as 9.17\% and R² increasing by as much as 35.99\%.
These findings indicate that RE'EM, and specifically, our constructed codebook, captures fundamental segregation mechanisms that extend beyond city-specific characteristics, making it a promising tool across diverse geographic and socio-economic settings.

\section{Discussion}

Our work highlights the transformative potential of LLMs in uncovering segregation experiences encoded in social media content. 
By revealing the nuanced and often invisible barriers that shape social interactions from online reviews, our work exemplifies the ethical and socially beneficial application of AI, demonstrating how advanced technologies can be harnessed to foster inclusivity and promote more equitable urban environments.
Policymakers, urban planners, and community leaders can leverage these insights to better understand and address patterns of social exclusion, ultimately working toward the development of more cohesive and diverse communities.

From a technical perspective, our framework advances the application of LLMs in tackling complex societal challenges. 
With reflective coding and integration of LLMs' reasoning and embedding capabilities, our framework not only significantly improves prediction accuracy but extends beyond conventional predictive tasks to offer explanations and actionable insights. 
Furthermore, our work demonstrates how prompt-guided analytical processes can mitigate biases and hallucinations, paving the way for more reliable and socially responsible AI applications.

Our work has several limitations.
Despite rigorous efforts to filter POIs with adequate review coverage, inherent imbalances in demographic representation across social media platforms may skew corpus distributions.
While our work demonstrates the overall effectiveness of using social media data to predict segregation, future work could further explore how and to what extent these data biases impact prediction accuracy across different POIs.
Besides, our analysis does not account for temporal variations in segregation, such as those driven by POI updates, policy shifts, or acute events like pandemics and economic crises. 
Future work could integrate temporal modeling to capture these dynamics.

\section{Conclusion}

In this paper, we pioneer the use of social media data to predict experienced segregation, designing a reflective LLM coder to generate insightful summaries and a REasoning-and-EMbedding (RE'EM) framework to integrate reasoning and embedding for accurate predictions. 
Our approach is validated through a qualitative user study and quantitative experiments, demonstrating its effectiveness in producing human-comprehensible review summaries and reliable segregation predictions. 
This work not only advances the understanding of experienced segregation but also provides a foundation for leveraging social media data to address broader societal challenges, such as fostering inclusivity and mitigating social inequalities. 

\section{Ethics Statement}
All datasets in this study were sourced from publicly available or academically licensed repositories, with privacy protections in place.
The review data was obtained from the Yelp Open Dataset under its Data Licensing agreement. 
The data was collected under users' consent and fully anonymized.
The Safegraph mobility data is currently accessible through the Dewey Data platform\footnote{\url{https://www.deweydata.io/}} for academic purposes.
This data was aggregated to the CBG level by month and processed with differential privacy techniques to safeguard against individual identity leakage. 
The ACS demographic data is publicly available\footnote{https://www.census.gov/programs-surveys/acs/}.
As such, no Institutional Review Board (IRB) approval was required by the authors’ institutions.


\begin{acks}
This work is supported in part by the National Natural Science Foundation of China (No.62472241) and the National Key Research and Development Program of China (2024YFC3307605).
\end{acks}

\bibliographystyle{ACM-Reference-Format}
\balance
\bibliography{sample-base}

@String{Computing = "Computing" }

@ArtifactSoftware{R,
    title = {R: A Language and Environment for Statistical Computing},
    author = {{R Core Team}},
    organization = {R Foundation for Statistical Computing},
    address = {Vienna, Austria},
    year = {2019},
    url = {https://www.R-project.org/},
}

@article{moro2021mobility,
  title={Mobility patterns are associated with experienced income segregation in large US cities},
  author={Moro, Esteban and Calacci, Dan and Dong, Xiaowen and Pentland, Alex},
  journal={Nature communications},
  volume={12},
  number={1},
  pages={4633},
  year={2021},
  publisher={Nature Publishing Group UK London}
}

@article{wang2018urban,
  title={Urban mobility and neighborhood isolation in America’s 50 largest cities},
  author={Wang, Qi and Phillips, Nolan Edward and Small, Mario L and Sampson, Robert J},
  journal={Proceedings of the National Academy of Sciences},
  volume={115},
  number={30},
  pages={7735--7740},
  year={2018},
  publisher={National Acad Sciences}
}

@article{tai2024examination,
  title={An examination of the use of large language models to aid analysis of textual data},
  author={Tai, Robert H and Bentley, Lillian R and Xia, Xin and Sitt, Jason M and Fankhauser, Sarah C and Chicas-Mosier, Ana M and Monteith, Barnas G},
  journal={International Journal of Qualitative Methods},
  volume={23},
  pages={16094069241231168},
  year={2024},
  publisher={SAGE Publications Sage CA: Los Angeles, CA}
}

@article{athey2021estimating,
  title={Estimating experienced racial segregation in US cities using large-scale GPS data},
  author={Athey, Susan and Ferguson, Billy and Gentzkow, Matthew and Schmidt, Tobias},
  journal={Proceedings of the National Academy of Sciences},
  volume={118},
  number={46},
  pages={e2026160118},
  year={2021},
  publisher={National Acad Sciences}
}

@article{de2024people,
  title={How people are exposed to neighborhoods racially different from their own},
  author={de la Prada, {\`A}lex G and Small, Mario L},
  journal={Proceedings of the National Academy of Sciences},
  volume={121},
  number={28},
  pages={e2401661121},
  year={2024},
  publisher={National Academy of Sciences}
}

@Book{park25city,
  author =       "Robert E. Park and Ernest W. Burgess",
  title =        "Understanding Policy-Based Networking",
  publisher =    "The University of Chicago Press",
  year =         "1925",
  address =      "Chicago, IL",
  edition =      "1st",
  editor =       "",
  volume =       "",
  number =       "",
  series =       "",
  month =        "",
  note =         "",
}

@article{charles2003dynamics,
  title={The dynamics of racial residential segregation},
  author={Charles, Camille Zubrinsky},
  journal={Annual review of sociology},
  volume={29},
  number={1},
  pages={167--207},
  year={2003},
  publisher={Annual Reviews 4139 El Camino Way, PO Box 10139, Palo Alto, CA 94303-0139, USA}
}

@article{massey1987effect,
  title={The effect of residential segregation on black social and economic well-being},
  author={Massey, Douglas S and Condran, Gretchen A and Denton, Nancy A},
  journal={Social Forces},
  volume={66},
  number={1},
  pages={29--56},
  year={1987},
  publisher={The University of North Carolina Press}
}

@article{king1973racial,
  title={Racial discrimination, segregation, and the price of housing},
  author={King, A Thomas and Mieszkowski, Peter},
  journal={Journal of political economy},
  volume={81},
  number={3},
  pages={590--606},
  year={1973},
  publisher={The University of Chicago Press}
}

@article{xu2016understanding,
  title={Understanding mobile traffic patterns of large scale cellular towers in urban environment},
  author={Xu, Fengli and Li, Yong and Wang, Huandong and Zhang, Pengyu and Jin, Depeng},
  journal={IEEE/ACM transactions on networking},
  volume={25},
  number={2},
  pages={1147--1161},
  year={2016},
  publisher={IEEE}
}

@article{yabe2023behavioral,
  title={Behavioral changes during the COVID-19 pandemic decreased income diversity of urban encounters},
  author={Yabe, Takahiro and Bueno, Bernardo Garc{\'\i}a Bulle and Dong, Xiaowen and Pentland, Alex and Moro, Esteban},
  journal={Nature communications},
  volume={14},
  number={1},
  pages={2310},
  year={2023},
  publisher={Nature Publishing Group UK London}
}

@inproceedings{chen2023getting,
  title={Getting Back on Track: Understanding COVID-19 Impact on Urban Mobility and Segregation with Location Service Data},
  author={Chen, Lin and Xu, Fengli and Hao, Qianyue and Hui, Pan and Li, Yong},
  booktitle={Proceedings of the International AAAI Conference on Web and Social Media},
  volume={17},
  pages={126--136},
  year={2023}
}

@inproceedings{luo2024othering,
  title={Othering and Low Status Framing of Immigrant Cuisines in US Restaurant Reviews and Large Language Models},
  author={Luo, Yiwei and Gligori{\'c}, Kristina and Jurafsky, Dan},
  booktitle={Proceedings of the International AAAI Conference on Web and Social Media},
  volume={18},
  pages={985--998},
  year={2024}
}

@inproceedings{iqbal2023lady,
  title={Lady and the Tramp Nextdoor: Online Manifestations of Real-World Inequalities in the Nextdoor Social Network},
  author={Iqbal, Waleed and Ghafouri, Vahid and Tyson, Gareth and Suarez-Tangil, Guillermo and Castro, Ignacio},
  booktitle={Proceedings of the International AAAI Conference on Web and Social Media},
  volume={17},
  pages={399--410},
  year={2023}
}

@inproceedings{santos2024real,
  title={REAL-UP: Urban Perceptions From LBSNs Helping Moving Real-Estate Market to the Next Level},
  author={Santos, Frances A and Silva, Thiago H and Villas, Leandro A},
  booktitle={Companion Proceedings of the ACM on Web Conference 2024},
  pages={1071--1074},
  year={2024}
}

@article{nguyen2016building,
  title={Building a national neighborhood dataset from geotagged Twitter data for indicators of happiness, diet, and physical activity},
  author={Nguyen, Quynh C and Li, Dapeng and Meng, Hsien-Wen and Kath, Suraj and Nsoesie, Elaine and Li, Feifei and Wen, Ming},
  journal={JMIR public health and surveillance},
  volume={2},
  number={2},
  pages={e5869},
  year={2016},
  publisher={JMIR Publications Inc., Toronto, Canada}
}

@article{mitchell2013geography,
  title={The geography of happiness: Connecting twitter sentiment and expression, demographics, and objective characteristics of place},
  author={Mitchell, Lewis and Frank, Morgan R and Harris, Kameron Decker and Dodds, Peter Sheridan and Danforth, Christopher M},
  journal={PloS one},
  volume={8},
  number={5},
  pages={e64417},
  year={2013},
  publisher={Public Library of Science San Francisco, USA}
}

@article{fatehkia2019correlated,
  title={Correlated impulses: Using Facebook interests to improve predictions of crime rates in urban areas},
  author={Fatehkia, Masoomali and O’Brien, Dan and Weber, Ingmar},
  journal={PloS one},
  volume={14},
  number={2},
  pages={e0211350},
  year={2019},
  publisher={Public Library of Science San Francisco, CA USA}
}

@inproceedings{rama2020facebook,
  title={Facebook ads as a demographic tool to measure the urban-rural divide},
  author={Rama, Daniele and Mejova, Yelena and Tizzoni, Michele and Kalimeri, Kyriaki and Weber, Ingmar},
  booktitle={Proceedings of The Web Conference 2020},
  pages={327--338},
  year={2020}
}

@article{li2024toward,
  title={Toward satisfactory public accessibility: A crowdsourcing approach through online reviews to inclusive urban design},
  author={Li, Lingyao and Hu, Songhua and Dai, Yinpei and Deng, Min and Momeni, Parisa and Laverghetta, Gabriel and Fan, Lizhou and Ma, Zihui and Wang, Xi and Ma, Siyuan and others},
  journal={arXiv preprint arXiv:2409.08459},
  year={2024}
}

@inproceedings{dai2023llm,
  title={LLM-in-the-loop: Leveraging Large Language Model for Thematic Analysis},
  author={Dai, Shih-Chieh and Xiong, Aiping and Ku, Lun-Wei},
  booktitle={Findings of the Association for Computational Linguistics: EMNLP 2023},
  pages={9993--10001},
  year={2023}
}

@inproceedings{gilbert2013need,
  title={" I need to try this"? a statistical overview of pinterest},
  author={Gilbert, Eric and Bakhshi, Saeideh and Chang, Shuo and Terveen, Loren},
  booktitle={Proceedings of the SIGCHI conference on human factors in computing systems},
  pages={2427--2436},
  year={2013}
}

@article{stier2021evidence,
  title={Evidence and theory for lower rates of depression in larger US urban areas},
  author={Stier, Andrew J and Schertz, Kathryn E and Rim, Nak Won and Cardenas-Iniguez, Carlos and Lahey, Benjamin B and Bettencourt, Lu{\'\i}s MA and Berman, Marc G},
  journal={Proceedings of the National Academy of Sciences},
  volume={118},
  number={31},
  pages={e2022472118},
  year={2021},
  publisher={National Acad Sciences}
}

@article{weller2014twitter,
  title={Twitter and society: An introduction},
  author={Weller, Katrin and Bruns, Axel and Burgess, Jean and Mahrt, Merja and Puschmann, Cornelius},
  journal={Twitter and society [Digital Formations, Volume 89]},
  pages={xxix--xxxviii},
  year={2014},
  publisher={Peter Lang Publishing}
}

@article{nilforoshan2023human,
  title={Human mobility networks reveal increased segregation in large cities},
  author={Nilforoshan, Hamed and Looi, Wenli and Pierson, Emma and Villanueva, Blanca and Fishman, Nic and Chen, Yiling and Sholar, John and Redbird, Beth and Grusky, David and Leskovec, Jure},
  journal={Nature},
  volume={624},
  number={7992},
  pages={586--592},
  year={2023},
  publisher={Nature Publishing Group UK London}
}

@article{wei2022chain,
  title={Chain-of-thought prompting elicits reasoning in large language models},
  author={Wei, Jason and Wang, Xuezhi and Schuurmans, Dale and Bosma, Maarten and Xia, Fei and Chi, Ed and Le, Quoc V and Zhou, Denny and others},
  journal={Advances in neural information processing systems},
  volume={35},
  pages={24824--24837},
  year={2022}
}

@article{GTE-base,
  title={Towards general text embeddings with multi-stage contrastive learning},
  author={Li, Zehan and Zhang, Xin and Zhang, Yanzhao and Long, Dingkun and Xie, Pengjun and Zhang, Meishan},
  journal={arXiv preprint arXiv:2308.03281},
  year={2023}
}

@inproceedings{BERT,
  title={BERT: Pre-training of Deep Bidirectional Transformers for Language Understanding},
  author={Jacob Devlin and Ming-Wei Chang and Kenton Lee and Kristina Toutanova},
  booktitle={North American Chapter of the Association for Computational Linguistics},
  year={2019},
  url={https://api.semanticscholar.org/CorpusID:52967399}
}

@inproceedings{glove,
	address = {Doha, Qatar},
	title = {{GloVe}: {Global} {Vectors} for {Word} {Representation}},
	url = {https://aclanthology.org/D14-1162/},
	doi = {10.3115/v1/D14-1162},
	booktitle = {Proceedings of the 2014 {Conference} on {Empirical} {Methods} in {Natural} {Language} {Processing} ({EMNLP})},
	publisher = {Association for Computational Linguistics},
	author = {Pennington, Jeffrey and Socher, Richard and Manning, Christopher},
	editor = {Moschitti, Alessandro and Pang, Bo and Daelemans, Walter},
	month = oct,
	year = {2014},
	pages = {1532--1543},
}

@article{GAT,
  title={Graph Attention Networks},
  author={Petar Velickovic and Guillem Cucurull and Arantxa Casanova and Adriana Romero and Pietro Lio’ and Yoshua Bengio},
  journal={ArXiv},
  year={2017},
  volume={abs/1710.10903},
  url={https://api.semanticscholar.org/CorpusID:3292002}
}

@article{gordon2006urban,
  title={Urban size, spatial segregation and inequality in educational outcomes},
  author={Gordon, Ian and Monastiriotis, Vassilis},
  journal={Urban Studies},
  volume={43},
  number={1},
  pages={213--236},
  year={2006},
  publisher={Sage Publications Sage UK: London, England}
}

@article{kramer2009segregation,
  title={Is segregation bad for your health?},
  author={Kramer, Michael R and Hogue, Carol R},
  journal={Epidemiologic reviews},
  volume={31},
  number={1},
  pages={178--194},
  year={2009},
  publisher={Oxford University Press}
}

@book{schelling2006micromotives,
  title={Micromotives and macrobehavior},
  author={Schelling, Thomas C},
  year={2006},
  publisher={WW Norton \& Company}
}

@article{chen2024shorter,
  title={Shorter Is Different: Characterizing the Dynamics of Short-Form Video Platforms},
  author={Chen, Zhilong and Liu, Peijie and Piao, Jinghua and Xu, Fengli and Li, Yong},
  journal={arXiv preprint arXiv:2410.16058},
  year={2024}
}

@article{qwen3embedding,
  title={Qwen3 Embedding: Advancing Text Embedding and Reranking Through Foundation Models},
  author={Zhang, Yanzhao and Li, Mingxin and Long, Dingkun and Zhang, Xin and Lin, Huan and Yang, Baosong and Xie, Pengjun and Yang, An and Liu, Dayiheng and Lin, Junyang and Huang, Fei and Zhou, Jingren},
  journal={arXiv preprint arXiv:2506.05176},
  year={2025}
}

@article{qwen3,
    title={Qwen3 Technical Report}, 
    author={An Yang and Anfeng Li and Baosong Yang and Beichen Zhang and Binyuan Hui and Bo Zheng and Bowen Yu and Chang Gao and Chengen Huang and Chenxu Lv and Chujie Zheng and Dayiheng Liu and Fan Zhou and Fei Huang and Feng Hu and Hao Ge and Haoran Wei and Huan Lin and Jialong Tang and Jian Yang and Jianhong Tu and Jianwei Zhang and Jianxin Yang and Jiaxi Yang and Jing Zhou and Jingren Zhou and Junyang Lin and Kai Dang and Keqin Bao and Kexin Yang and Le Yu and Lianghao Deng and Mei Li and Mingfeng Xue and Mingze Li and Pei Zhang and Peng Wang and Qin Zhu and Rui Men and Ruize Gao and Shixuan Liu and Shuang Luo and Tianhao Li and Tianyi Tang and Wenbiao Yin and Xingzhang Ren and Xinyu Wang and Xinyu Zhang and Xuancheng Ren and Yang Fan and Yang Su and Yichang Zhang and Yinger Zhang and Yu Wan and Yuqiong Liu and Zekun Wang and Zeyu Cui and Zhenru Zhang and Zhipeng Zhou and Zihan Qiu},
    journal = {arXiv preprint arXiv:2505.09388},
    year={2025}
}

@misc{meta2025llama4,
  author       = {Meta AI},
  title        = {llama4},
  year         = {2025},
  howpublished={\url{https://www.llama.com/models/llama-4/}}
}

@misc{deepseek2025v32exp,
      title={DeepSeek-V3.2-Exp: Boosting Long-Context Efficiency with DeepSeek Sparse Attention}, 
      author={DeepSeek-AI},
      year={2025},
}

\clearpage
\appendix

\section{Variability of POI features}
\label{appendix:a}
\begin{figure}[ht]
    \centering
    \includegraphics[width=0.85\linewidth]{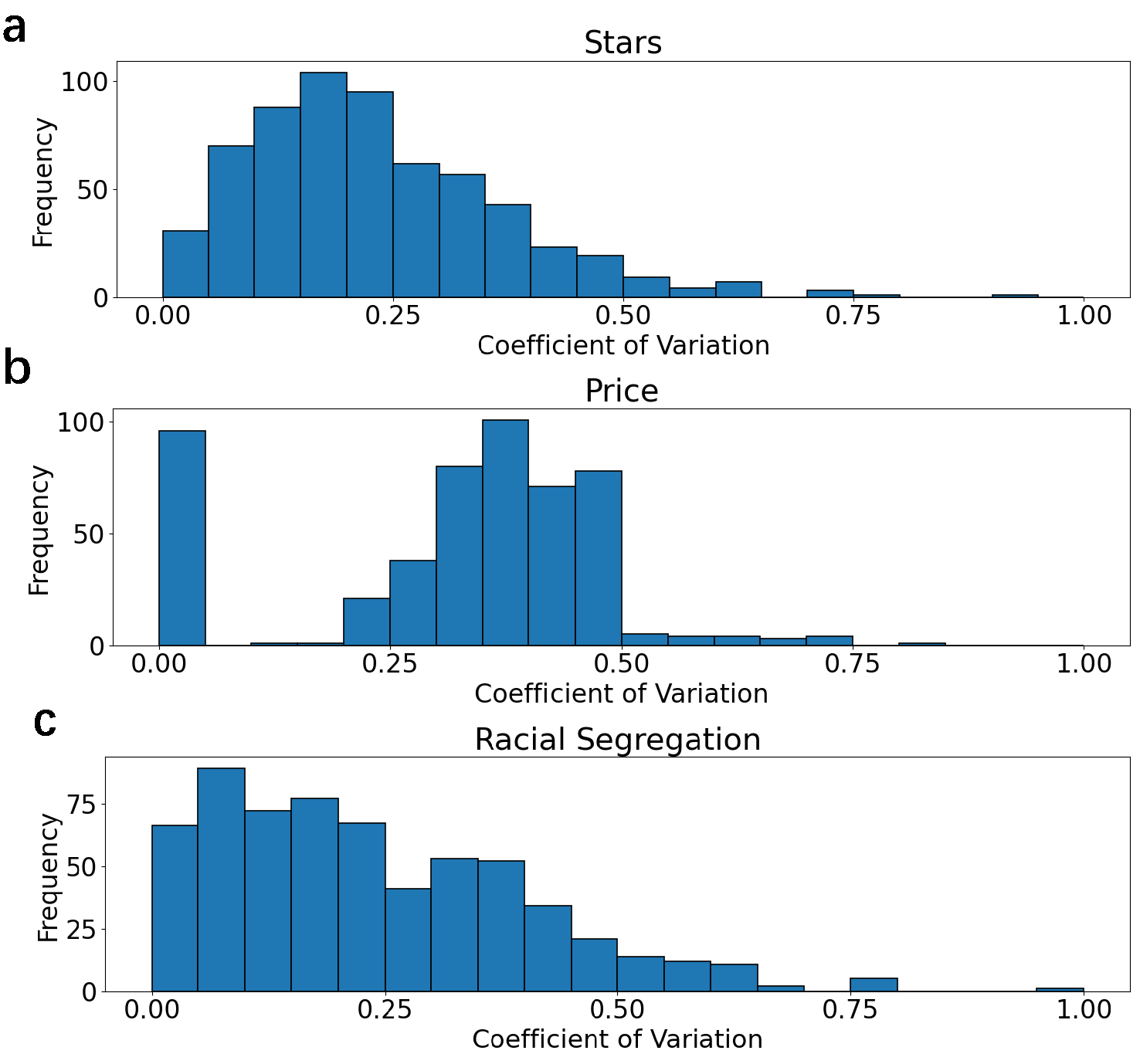}
    \caption{The Coefficient of Variation distributions for POIs features within the same CBG highlight considerable variability among POIs. Figures a-c demonstrate the distributions of stars, price range, and racial segregation, respectively. 
    }
    \label{fig:poi-variation-statistics}
\end{figure}

Taking Philadelphia as an example, Figure~\ref{fig:poi-variation-statistics} illustrates the distributions of the variations for POI features within a single CBG, including stars, prices, income segregation, and racial segregation.
We can observe that POIs exhibit considerable variability even when examined within the CBG locality, which underscores the significance of our study.

\section{Reflective Coding Algorithm}
We formalize the process of reflective coding in Algorithm~\ref{algorithm}.
\label{appendix:b}

\section{Implementation Details}
\label{appendix:c}

To obtain a representative subset for the coding process, we categorize all POIs into 36 types base on different combinations of visitor-residential composition gaps (e.g., higher \textit{black} and \textit{asian} ratios with lower \textit{white}, \textit{hispanic}, and \textit{other} ratios) and perform stratified sampling within each type to ensure diversity.

For the \textit{reasoning} channel, we train an MLP with hidden dimensions of 512, 128, and 64, removing the output layer to form the Rating Adapter.
For the \textit{embedding} channel, we fine-tune the last two layers of the pre-trained GTE-base model and extend it with three fully connected layers (512, 256, 128) to form the Embedding Adapter.
For the \textit{population} channel, we train an MLP with dimensions of 100, 30, and 10, extracting the weights of the first two layers to form the Population Adapter.
For the \textit{neighbor-aware multi-view predictor}, we adopt Graph Attention Networks (GATs)~\cite{GAT} as the neighbor aggregators. 
The Reasoning GAT maps feature from 64 to 128, the Embedding GAT maintains a 128 to 128 mapping, and the Population GAT maintains a 30 to 30 mapping. 
The multi-view fusion module comprises three fully connected layers with dimensions 512, 128, and 64. 
During training, we first optimize the three Adapters, then freeze them before training the predictor.

We conduct a compact grid search on the Philadelphia dataset, exploring learning rates in $[5\times10^{-6}, 5\times10^{-3}]$, weight decays in $[10^{-5}, 5\times10^{-4}]$, early-stopping patience between 10 and 25.
The width of every MLP hidden layer is tuned by doubling or halving around the final choices (approximately $64$–$512$ units).  
The configuration achieving the lowest validation MSE is selected and is released as the default value in our open-source code.

All experiments run on a Linux server with an Intel Xeon Platinum 8358 CPU, 4$\times$RTX 4090 and 2$\times$RTX 3090 GPUs (CUDA 12.4), and 503~GB RAM.



\begin{figure}[ht]
\begin{algorithm}[H]
\caption{Reflective Coding Algorithm}
\label{algorithm}

\begin{algorithmic}[1]
\Require Set of POIs $\mathcal{I}$ with for each POI $i \in \mathcal{I}$:
\Statex \quad - Social media content $c_i = (\mathcal{N}_i, \{\text{review}_k, \text{image}_k\}_{k=1}^{m_i})$
\Statex \quad - Observed visitor composition $\boldsymbol{\tau}_i = [\tau_{i q}]_{q=1}^Q$ (Q racial groups)
\Statex \quad - Local residential composition $\mathbf{p}_i = [p_{i q}]_{q=1}^Q$

\Ensure Codebook $\mathcal{B} = \{(d_j, e_j)\}_{j=1}^M$ with dimensions $d_j$ and explanations $e_j$

\State \textbf{// Phase 1: Reflective Attributor}
\State Initialize insight set $\mathcal{L} \gets \emptyset$

\For{each POI $i \in \mathcal{I}$}
    \State \textbf{// Multi-source insight generation}
    \State $\mathcal{L}_i^{\text{name}} \gets \text{LLM}_\text{analyze}(\mathcal{N}_i)$ \Comment{Extract name signals}
    \State $\mathcal{L}_i^{\text{text}} \gets \text{LLM}_\text{analyze}(\{\text{review}_k\})$ \Comment{Analyze textual reviews}
    \State $\mathcal{L}_i^{\text{img}} \gets \text{LLM}_\text{analyze}(\{\text{image}_k\})$ \Comment{Analyze visual content}
    \State $\mathcal{L}_i^{\text{sum}} \gets \text{LLM}_\text{synthesize}(\mathcal{L}_i^{\text{name}}, \mathcal{L}_i^{\text{text}}, \mathcal{L}_i^{\text{img}})$
    
    \State \textbf{// Reflection-guided refinement via abductive reasoning}
    \State $\Delta\boldsymbol{\tau}_i \gets \boldsymbol{\tau}_i - \mathbf{p}_i$ \Comment{Compute demographic discrepancy}
    \State $\mathcal{L}_i^{\text{refined}} \gets \text{LLM}_\text{reflect}(\mathcal{L}_i^{\text{sum}}, \Delta\boldsymbol{\tau}_i)$
    \State $\mathcal{L}[i] \gets \mathcal{L}_i^{\text{refined}}$ 
\EndFor

\State \textbf{// Phase 2: Code Summarizer}
\State Partition $\mathcal{L}$ into $K$ stratified subsets: $\{\mathcal{L}_{\text{sub}}^{(k)}\}_{k=1}^K$
\State Initialize codebook $\mathcal{B} \gets \emptyset$

\For{each subset $\mathcal{L}_{\text{sub}}$}
    \State \textbf{// Extract candidate codes}
    \State $\mathcal{C}_{\text{sub}} \gets \emptyset$
    \For{each insight $\ell \in \mathcal{L}_{\text{sub}}$}
        \State $(c, e) \gets \text{LLM}_\text{extract}(\ell)$
        \State $\mathcal{C}_{\text{sub}} \gets \mathcal{C}_{\text{sub}} \cup \{(c, e)\}$
    \EndFor
    
    \State \textbf{// Iterative refinement on combined set}
    \State $\mathcal{B} \gets \mathcal{B} \cup \mathcal{C}_{\text{sub}}$ \Comment{Merge candidates with existing codebook}
    \State $\mathcal{B} \gets \text{LLM}_\text{semantic}(\mathcal{B})$ \Comment{LLM merges redundant dimensions}
    \State $\mathcal{B} \gets \text{LLM}_\text{quality}(\mathcal{B})$ \Comment{LLM filters low-generalizability codes}
    \State $\mathcal{B} \gets \text{LLM}_\text{refine}(\mathcal{B})$ \Comment{LLM generates thematic descriptions}
\EndFor

\State \Return $\mathcal{B}$
\end{algorithmic}
\end{algorithm}
\end{figure}


\begin{table*}[ht]
    \centering
    \caption{Evaluation of different base LLMs in the RE’EM framework. Replacing GPT-4o-mini with other SOTA LLMs yields nearly identical results, showing that RE'EM is robust to the choice of base model.}
    \vspace{-10pt}
    \scalebox{0.99}{
    \begin{tabular}{@{}ccccc@{}}
    \toprule
    Base LLM & MSE↓ & RMSE↓ & MAE↓ & R$^2$↑ \\ \midrule
GPT-4o-mini      & 0.0068±0.0003 & 0.0823±0.0021 & 0.0662±0.0021 & 0.3885±0.0278 \\
Qwen-plus        & 0.0068±0.0003 & 0.0822±0.0021 & 0.0661±0.0021 & 0.3905±0.0289 \\
Llama4-scout     & 0.0068±0.0004 & 0.0823±0.0022 & 0.0660±0.0021 & 0.3896±0.0305 \\
DeepSeek-v3.2-exp& 0.0068±0.0004 & 0.0823±0.0022 & 0.0662±0.0021 & 0.3883±0.0301 \\
\bottomrule
    \end{tabular}
    }
    \label{tab:LLMBase}
\end{table*}

\begin{table*}[ht]
    \centering
    \caption{Comparison with direct LLM prompting baseline. Removing RE'EM pipeline leads to severe performance degradation, confirming the necessity of RE'EM.}
    \vspace{-10pt}
    \scalebox{0.99}{
    \begin{tabular}{@{}ccccc@{}}
    \toprule
    Model & MSE↓ & RMSE↓ & MAE↓ & R$^2$↑ \\ \midrule
Qwen-plus & 0.1281±0.0039 & 0.3579±0.0054 & 0.3202±0.0068 & -10.7420±0.4472 \\
Llama4-scout & 0.0586±0.0020 & 0.2420±0.0043 & 0.1916±0.0042 & -4.3700±0.2091 \\
DeepSeek-v3.2-exp & 0.0689±0.0022 & 0.2625±0.0042 & 0.2127±0.0030 & -5.3191±0.2724 \\
RE’EM              & \textbf{0.0068±0.0003} & \textbf{0.0823±0.0021} & \textbf{0.0662±0.0021} & \textbf{0.3885±0.0278} \\
\bottomrule
    \end{tabular}
    }
    \label{tab:DirectLLM}
\end{table*}

\begin{figure*}[ht]
    \centering
    \includegraphics[width=0.95\linewidth]{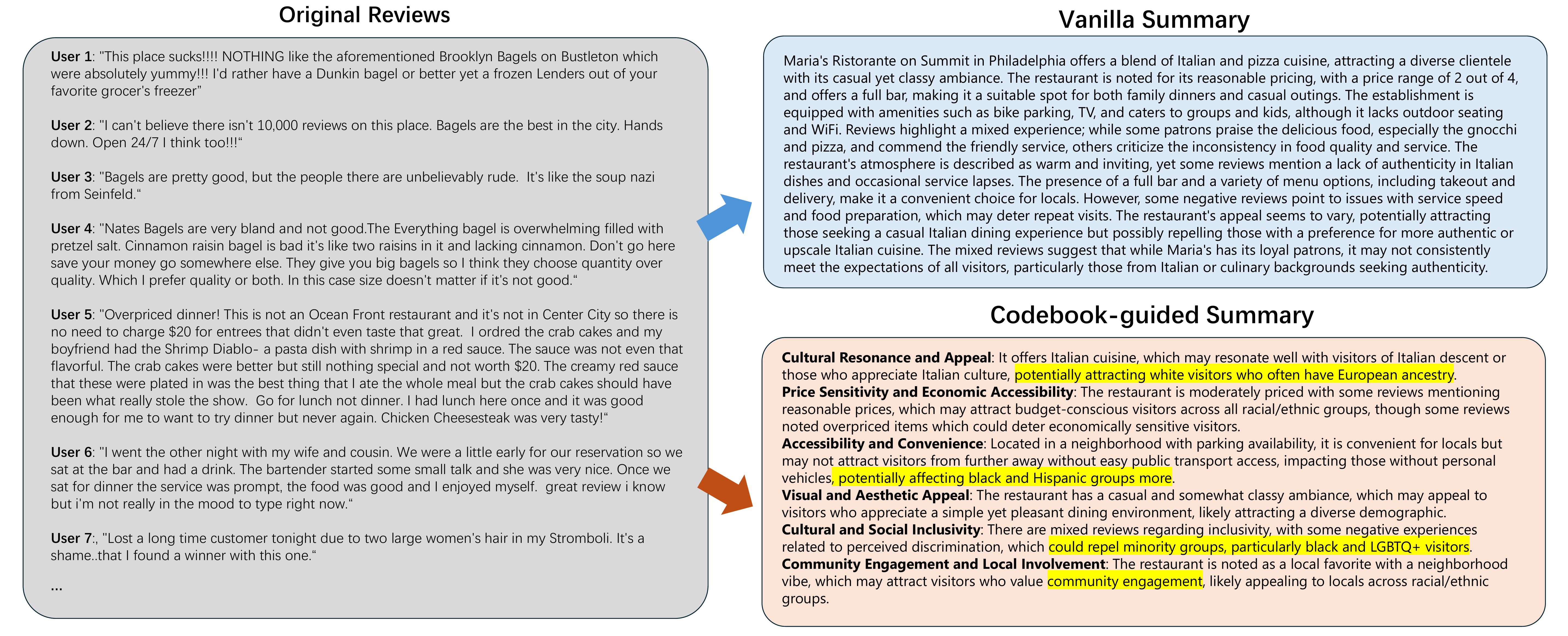}
    \caption{Case study comparing the vanilla summary and codebook-guided summary for the same venue. The codebook-guided summary captures more nuanced factors influencing demographic appeal, which are absent in the vanilla summary, illustrating the interpretability and analytical advantage of structured reasoning.}
    \label{fig:case-study}
\end{figure*}

\section{Additional Evaluation on Different LLMs}
\label{appendix:additional}
To further examine whether RE’EM depends on a particular language model, we conduct two additional evaluations involving both open- and closed-source state-of-the-art LLMs. 
First, we replace the base model (GPT-4o-mini) used in the reasoning channel with three strong alternatives: Qwen-plus~\cite{qwen3}, Llama4-scout~\cite{meta2025llama4}, and DeepSeek-v3.2-exp~\cite{deepseek2025v32exp}, keeping all other components unchanged. 
As shown in Table~\ref{tab:LLMBase}, the performance across all metrics remains nearly identical, indicating that RE'EM is highly robust to the choice of the base LLM.

Second, we assess whether the performance gain comes from the RE'EM architecture rather than the inherent ability of LLMs. 
We directly prompt the same LLMs to predict segregation from social media content and population distribution, removing the RE'EM pipeline. 
Table~\ref{tab:DirectLLM} shows a dramatic drop in performance for all models, confirming that naive LLM prompting is insufficient and that the structured multi-channel design of RE’EM is essential for reliable quantitative prediction.

\section{Case study}
\label{appendix:d}
In Figure~\ref{fig:case-study}, we present a case study to compare the original reviews, the vanilla method-generated summary, and our codebook-guided summary.



\end{document}